\documentclass{article}

\usepackage{PRIMEarxiv}

\usepackage[utf8]{inputenc} 
\usepackage[T1]{fontenc}    
\usepackage{hyperref}       
\usepackage{url}            
\usepackage{booktabs}       
\usepackage{amsfonts}       
\usepackage{nicefrac}       
\usepackage{microtype}      
\usepackage{lipsum}
\usepackage{fancyhdr}       
\usepackage{graphicx}       
\graphicspath{{media/}}     

\usepackage{caption}
\usepackage{subcaption}
\usepackage{endnotes}

\title{Transformer-based Named Entity Recognition with Combined Data Representation}

\author{
  Michał Marcińczuk \\
  CodeNLP \\
  \texttt{marcinczuk@gmail.com} \\
}

\begin{document}
\maketitle

\begin{abstract}
This study examines transformer-based models and their effectiveness in named entity recognition tasks. The study investigates data representation strategies, including single, merged, and context, which respectively use one sentence, multiple sentences, and sentences joined with attention to context per vector. Analysis shows that training models with a single strategy may lead to poor performance on different data representations. To address this limitation, the study proposes a combined training procedure that utilizes all three strategies to improve model stability and adaptability. The results of this approach are presented and discussed for four languages (English, Polish, Czech, and German) across various datasets, demonstrating the effectiveness of the combined strategy.
\end{abstract}


\section{Introduction}
Identifying specific word sequences that belong to certain entity categories, such as people, locations, organizations, objects, and events, is the main goal of named entity recognition (NER) in natural language processing. However, the definition of a named entity differs in each application and dataset, which creates challenges for NER. 
The recognition of named entities poses several challenges. One challenge arises because many named entities are proper names, which are rigid designators \cite{Kripke1980-KRINAN}. Proper names serve as references to entities and do not inherently provide descriptive information about those entities. Consequently, identifying named entities requires prior knowledge that a particular term qualifies as a named entity. Furthermore, the set of possible named entities is vast and limitless. Several approaches can be used to train a model to recognize a term as a named entity. These include training on annotated datasets that identify named entities, incorporating a common sense knowledge base about the world, or inferring characteristics from the training data. For example, consider the sentence "Mark works in Xax." The term "Xax" is probably a named entity because it is capitalized. However, it may require additional information to determine its semantic category, such as whether it refers to a location (city or country) or a company name. Subsequent sentences like "He loves this city." could provide contextual clues, clarifying that "Xax" is indeed the name of a city. It is worth noting that the less information available from the input text, the greater the reliance on information contained in the training dataset or external sources. Current state-of-the-art methods in the field of named entity recognition are based on pre-trained language models in a transformers architecture \cite{ye2021pack,wang-etal-2021-improving,yamada2020luke,contextBert2020}. Transformer-based models outperform approaches like long short-term memory (LSTM) networks or conditional random fields (CRF) \cite{poldeepner2018}.

Our research focuses on different data representations used during training and inference for transformer-based models and their impact on performance. The pioneering application of a transformer-based model to the NER task presented by Devlin et al. (2019) \cite{devlin2019bert} used a maximal document context provided by the data. Jouni and Sampo \cite{luoma-pyysalo-2020-exploring} (2020) and Wang et al.~\cite{wang-etal-2021-improving} (2021) showed that context-based representation produces better results than processing single sentences. This research shows a positive impact of long dependencies between text fragments. However, we discovered that a model trained solely on data with context loses performance on shorter text fragments. This occurs when the model receives single sentences instead of whole documents for inference. At the same time, a model trained on single sentences performs significantly better than the one trained with context when processing single sentences. 

The contribution of our research is as follows:
\begin{itemize}
    \item evaluation of NER models for various data representations during training and inference on five publicly available NER datasets,
    \item proposal of a training technique that makes the model more resilient on data representations on inference. 
\end{itemize}

The paper is organized as follows. Section 2 presents a review of recent trends in NER. In Section 3, we describe the research methodology of the NER task, including the data representations, datasets used in the experiments, parameter values used for training the models, and metrics used for model comparison. In Section 4, we discuss the results obtained in the initial experiments that formed the basis of our research, and highlight the performance differences for different data representations. Section 5 introduces and evaluates the improved training procedure. Section 6 compares the obtained results with the state-of-the-art. Finally, in Section 7, we compare our results with the current state-of-the-art for the used corpora.

\section{Related work}

The current state-of-the-art approaches for NER are based on transformer-based pre-trained language models. The first model that utilized this architecture was BERT, presented by Devlin, et al. in their research paper \cite{devlin2019bert}. Since then, there have been several directions in which NER models have been developed. One approach is the utilization of better language models. Liu, et al. \cite{Liu2019RoBERTaAR} presented an optimized and more robust technique for BERT pre-training. Chen, et al. (2022) \cite{chen-etal-2022-comparative} showed that different variants of pre-trained models obtain various results on different datasets, and there is no one model that is superior over the other. The performance depends on the NER dataset used for training. Wang, et al. (2021) \cite{DBLP:journals/corr/abs-2010-05006} showed that concatenating embeddings from multiple language models can further improve the performance, but at the cost of increased computational complexity.

The development of NER is progressing in two directions: utilizing external knowledge with pre-trained language models and improving model performance with document-level contexts. Yamada et al. (2020) \cite{yamada2020luke} proposed a model that uses pre-trained contextualized representations of words and entities based on a bidirectional transformer, along with an entity-aware self-attention mechanism. The model was pre-trained on entities from a large entity-annotated corpus retrieved from Wikipedia. Wang et al. (2021) \cite{wang-etal-2021-improving} showed that adding document-level contexts can significantly improve model performance. They proposed a technique that extends the original text with external contexts of a sentence by retrieving and selecting a set of semantically relevant texts through a search engine, using the original sentence as the query. This approach was utilized extensively in the SemEval-2023 Task 2: MultiCoNER 2 shared-task, which focused on processing short and noisy text snippets. The winning submission by Tan et al. (2023) \cite{tan-etal-2023-damo} introduced a unified retrieval-augmented system that fetches and ranks possible context from external sources.

In the early months of 2023, the popularity of large language models (LLMs) rose significantly, mainly due to the ChatGPT, GPT-4 and LLaMa models. Various researchers have demonstrated the impressive zero-shot performance of these models across a wide range of tasks (Laskar et al., 2023, \cite{laskar-etal-2023-systematic}). However, Wang et al. (2023) \cite{wang2023gptner} conducted an extensive evaluation of named entity recognition (NER) using LLM models and found that their performance still falls significantly below the state-of-the-art standards. Li et al. (2023) \cite{Li2023PromptingCI} proposed a different way of utilizing LLM models like ChatGPT. The idea is to use ChatGPT as an implicit knowledge base and enable it to generate auxiliary knowledge heuristically for more efficient entity prediction.

\section{Methodology}

\subsection{Named entity recognition with transformers}

\begin{figure}[ht]
    \centering
    \includegraphics[width=0.8\textwidth]{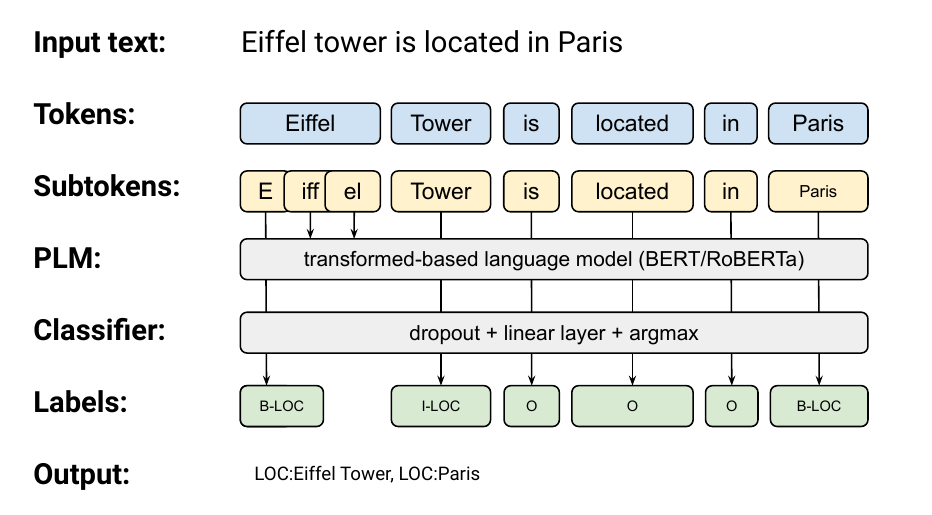}
    \caption{The neural network architecture for named entity recognition.}
    \label{fig:nn}
\end{figure}

To model the problem of named entity recognition with transformers, we follow the original approach presented by Devlin et al. (2019)~\cite{devlin2019bert}. This approach represents the recognition of named entities as a sequence classification task. The input to the model consists of a sequence of subtokens produced by the tokenizer. The model produces a sequence of labels representing the boundaries of named entities. The labels are aligned with words, meaning only the first subtoken of each word is used as the input to the token-level classifier over the label set. The named entity labels are encoded using the IOB2 schema (Inside, Outside, and Begin). We followed the procedure presented by \cite{MARCINCZUK2021291} to encode nested entities. Every nested entity is decoded separately using the IOB2 schema in this approach. Then, all the entity-level labels are concatenated for each token into a single word-level label. For instance, a word that begins a person's name and is inside an organization name will be encoded as \verb|I-ORG#B-PER|, where \verb|#| jest a label separator. 

The neural network architecture consists of two main elements: a pre-trained language model (PLM) embeddings and a classifier layer (see Figure~\ref{fig:nn}). The PLM part generates a context-based representation of each word (a vector of the first word's subtoken). The classifier layer outputs one of the IOB2 labels for each word. The label set depends on a given dataset's categories of named entities. In the post-processing, the sequences of IOB2 labels are wrapped into continuous annotations. The text is divided into tokens, and then each token is divided into subtokens by the model tokenizer. Some tokes are divided into more than one subtoken. For example, the \textit{Eiffel} word is tokenized into three subtokens: \textit{E}, \textit{iff}, and \textit{el}). In such a case, only the first subtoken is passed through the classifier to obtain the label of the word. This approach is sufficient due to the attention mechanism of transform-based models. 

The model processes the text in small batches, each containing a fixed number of subwords. The number of subwords depends on the architecture of PLM. The first PLMs, such as BERT~\cite{devlin2019bert} and RoBERTa~\cite{zhuang-etal-2021-robustly}, used a size of 512 subwords. There are also PLMs that offer much longer sequences. For instance, BigBird~\cite{zaheer2021big}, and Longformer~\cite{beltagy2020longformer} can handle sequences up to 4096 subtokens. The transformer architecture uses a multi-head attention mechanism, which allows it to learn long-range dependencies between subtokens in a sequence. Therefore, dividing the text into fragments is crucial for data processing. The goal is to divide the text into fragments that are small enough for the transformer to learn the long-range dependencies but large enough to capture the important information in the text. In the next section, we discuss different strategies of data representation.

\subsection{Data representation}
\label{sec:representations}

The transformer-based models take into account two factors for data representation --- split of document text (how the document is split into smaller pieces) \cite{devlin2019bert} and external context (out-of-document knowledge added to the document) \cite{tan-etal-2023-damo}. Our research focuses only on in-document representation. 

We deal with the problem of in-document representation when the tokenized text is longer than the maximum sequence length of the PLM (for example, 512 for the BERT base model). In such a case, the document should be divided into a list of shorter fragments. Then, the fragments are processed independently, and the outputs are combined into the final result. Below, we describe three data representation techniques.

\subsubsection{Single}

\begin{figure}[h]
     \centering
     \includegraphics[width=0.7\textwidth]{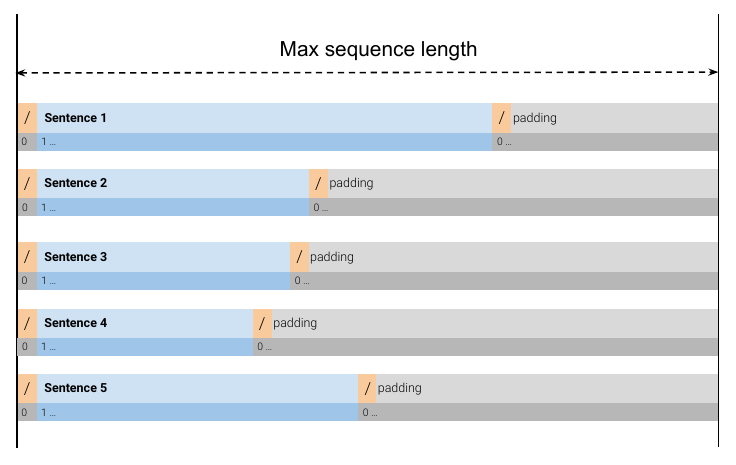}
     \caption{Visualization of the single sentence data representation. The blue parts represent the subtokens that the model will process. Grey parts are paddings that fill the vector to the maximum sequence size.}
     \label{fig:representation-single}
 \end{figure}

    Text is divided into sentences. Each vector contains exactly one sentence. If a sentence is longer than the maximum number of subtokens, it is split into several non-overlapping vectors. This data representation is visualized in Figure~\ref{fig:representation-single}. 
    
    In this representation, we limit the impact of long dependences on the representation of the sentence. This representation also applies while processing short texts (single sentences), as the text is already limited to a single sentence. 

\subsubsection{Merged}

\begin{figure}[h]
     \centering
     \includegraphics[width=0.7\textwidth]{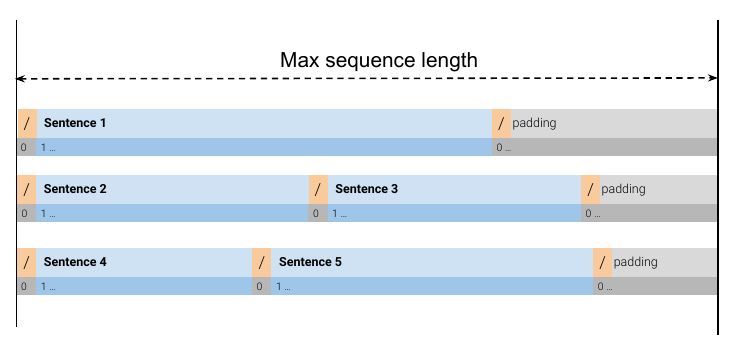}
     \caption{Visualization of the merged data representation. The blue parts represent the subtokens that the model will process. Grey parts are paddings that fill the vector to the maximum sequence size.}
     \label{fig:representation-merged}
 \end{figure}

    Text is divided into sentences. Each vector can contain more than one sentence. The vectors are filled with sentences until the limit of maximum subtokens is reached. This approach provides high-speed improvements, as it reduces the number of vectors to process. This data representation is visualized in Figure~\ref{fig:representation-merged}.

\subsubsection{Context}
    
\begin{figure}[h]
     \centering
     \includegraphics[width=0.7\textwidth]{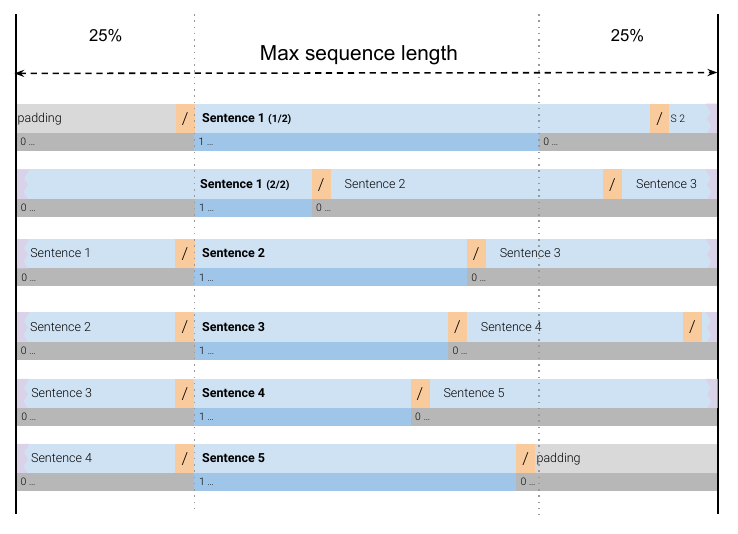}
     \caption{Visualization of the context data representation. The blue parts represent the subtokens that the model will process. Grey parts are paddings that fill the vector to the maximum sequence size. The 25\% is a sample size of the context width. Those parts are considered when calculating attention but are not passed through the classification layer.}
     \label{fig:representation-context}
 \end{figure}

    The document is divided into smaller fragments (sentences or fragments of fixed width). A vector is generated for every fragment. Each fragment is extended by left and right context by adding preceding and following fragments. The additional fragments serve as a context and are masked for the classification layers. This means that they are used only to calculate the attention but are not passed through the token classifier layer. This data representation is visualized in Figure~\ref{fig:representation-context}. The advantage of this representation is the efficient context usage for embedding generation with the attention mechanism. This strategy produced the best performance \cite{devlin2019bert,luoma-pyysalo-2020-exploring}.

\subsection{Datasets}

\begin{table}[ht]
    \caption{List of datasets used in the research.}
    \small
    \centering
    \begin{tabular}{lcrrrrc}
        \toprule
         \textbf{Corpus} & \textbf{Language} & \textbf{Sentences} & \textbf{Tokens} & \textbf{Annotations} & \textbf{Categories} & \textbf{Nested} \\ 
         \midrule
         CoNLL 2003       & English &  20~744 &   301~418 &  35~089  &  4  & No  \\
         GermEval 2014    & German  &  31~300 &   591~005 &  40~941  & 12  & Yes \\
         CNEC 2.0         & Czech   &   8~992 &   199~965 &  28~727  & 46  & Yes \\
         NKJP             & Polish  & 153~982 & 2~180~611 & 154~292  & 14  & Yes \\
         KPWr (n82)       & Polish  &  18~282 &   303~678 &  17~716  & 82  & Yes \\
        \bottomrule
    \end{tabular}
    \label{tab:datasets}
\end{table}

In our research, we used five publicly available NER datasets summarized in Table~\ref{tab:datasets}. With this selection of datasets, we cover the following aspects of NER task:
\begin{itemize}
    \item language --- English, Polish, German, and Czech,
    \item corpus size --- from 200k tokens for CNEC 2.0 to 2.2M for NKJP,
    \item number of entity categories --- from 4 for CoNLL 2003 to 82 for KPWr.
\end{itemize}

The following subsections contain a brief description of each dataset.

\subsubsection{CoNLL 2003 (English)}

CoNLL 2003 \footnote{\url{https://www.clips.uantwerpen.be/conll2003/ner/}} \cite{sang2003introduction} is a NER dataset for English. It consists of news articles from various sources. The dataset contains 35k named entities of four coarse-grained categories of entities: person, location, organization, and miscellaneous. 

\subsubsection{GermEval 2014 (German)}

GermEval 2014\footnote{\url{https://sites.google.com/site/germeval2014ner/data}} \cite{BenikovaBiemannKisselewetal.2014} is a NER dataset for German. The dataset consists of samples from German Wikipedia and News Corpora. The dataset contains almost 41k named entities of four (person, location, organization, and other) coarse-grained categories and eight subcategories (\textit{deriv} and \textit{part} for each coarse-grained category). This makes a total of 12 categories of named entities. The annotation schema defines a two-level nesting of entities.

\subsubsection{CNEC 2.0 (Czech)}

\begin{figure}[h]
    \centering
    \includegraphics[width=0.8\textwidth]{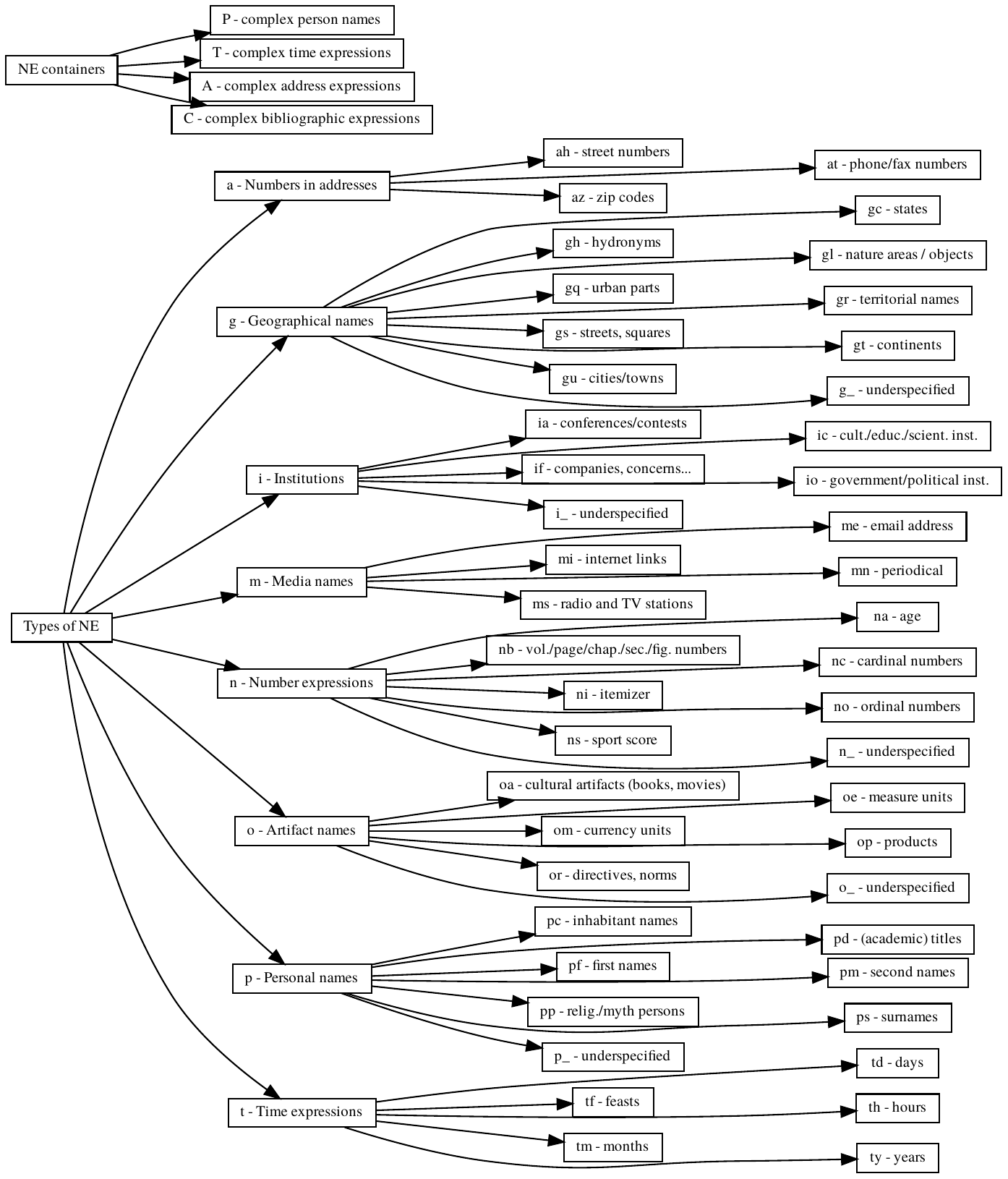}
    \caption{Schema of named entity categories in the CNEC 2.0 dataset.}
    \label{fig:cnec}
\end{figure}

CNEC 2.0\footnote{\url{https://ufal.mff.cuni.cz/cnec/cnec2.0}} \cite{11858/00-097C-0000-0023-1B22-8} is a NER dataset for Czech. The dataset contains 28.7k annotations classified according to a two-level hierarchy of 46 fine-grained categories (see Figure~\ref{fig:cnec}). Annotations have up to four levels of nesting.

\subsubsection{NKJP (Polish)}

\begin{figure}[h!]
    \centering
    \includegraphics[width=0.8\textwidth]{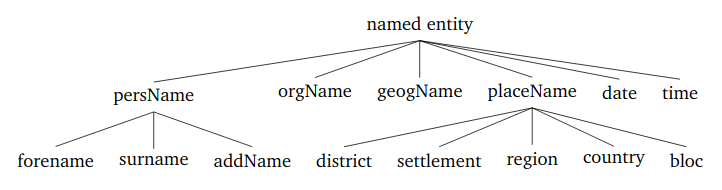}
    \caption{Schema of named entity categories in the NKJP dataset.}
    \label{fig:nkjp}
\end{figure}

NKJP (The National Corpus of Polish)\footnote{\url{https://nkjp.pl/?lang=1}} \cite{NKJP} is a collection of texts of various genres, including classic literature, daily newspapers, specialist periodicals and journals, conversation transcripts, and various short-lived and Internet texts. The dataset contains 154k annotations of 14 categories of named entities (see Figure~\ref{fig:nkjp}). Annotations have multiple levels of nesting.

The dataset has no official split into train, dev, and test subsets. Thus, we used our custom division. We used 80\% of the dataset as a train and the remaining 20\% as a tune subset.



\subsubsection{KPWr (Polish)}

KPWr\footnote{\url{https://clarin-pl.eu/dspace/handle/11321/722}} \cite{g419-kpwr-2012} is a collection of Polish texts available under an open license. It contains more than 1~400 short excerpts from texts of various genres with 17.7 annotations. The annotation schema defines 82 fine-grained categories of named entities. 

\subsection{Training parameters}

For each experiment, we used the same set of parameters. The only difference was the PLM, which depends on the language. The parameters are summarized in Table~\ref{tab:parameters}.

\begin{table}[h!]
    \caption{Training parameters of the neural network}
    \small
    \centering
    \begin{tabular}{ll}
        \toprule
        \textbf{Parameter} & \textbf{Value} \\
        \midrule
            Sequence length & 256 \\
            Dropout         & 0.2 \\
            Epochs          & 20 \\
            Learning rate & 5e-6 \\
            PLM & - \emph{xlm-roberta-large} for English, Czech, and German \\
                & - \emph{allegro/herbert-large-cased} for Polish \\
            Optimizer & AdamW \\
            Scheduler & Liner without warmup \\
        \bottomrule
    \end{tabular}
    \label{tab:parameters}
\end{table}

\subsection{Metrics}

We report an average F1 score and standard deviation ($\sigma$) from five runs for every experiment configuration. The F1 score is calculated for precision and recall using strict matching on the level of annotation spans. We follow the metrics used on the CoNLL 2003 NER task \cite{sang2003introduction}.

\section{Initial results and observations}

\subsection{Introduction}

In this section, we present the initial results that form the basis of our research. We conducted experiments to evaluate the impact of different data representations during both training and inference on five datasets. This resulted in nine different combinations and a total of 45 configurations. We conducted all experiments, even those that had already been reported (such as Goyal et al., 2020~\cite{conneau-etal-2020-unsupervised} results on the CoNNL 2003 dataset), to ensure a fair comparison and to isolate data representation as the sole variable. All results are reported for the tune parts of the datasets. For a detailed presentation of the results, please refer to Appendix~\ref{sec:app1}.

\subsection{Performance discrepancy between strategies}
\label{sec:discrapency}

We trained three models for each dataset using three different data representation techniques: single, merged, and context. Each model was evaluated on the tune subset using the same data representation for inference as was used for training. The results are presented in Table~\ref{tab:models_strategy_eval}. The context representation led to the best performance for each dataset. The difference between the second-best score varied from 0.3 percentage points (pp) for GermEval 2014 to 16.48 pp for KPWr (n82). This confirms previous research by Devlin et al. (2019)~\cite{devlin2019bert} and Luoma-Pyysalo (2020)~\cite{luoma-pyysalo-2020-exploring} that document-level context improves the performance of NER.

\begin{table}[ht]
    \caption{An average of F1 from 5 runs on the tune parts of the datasets using the same setup for training and testing.}
    \centering
    \begin{tabular}{lcccccc}
        \toprule
         \textbf{Corpus} & \multicolumn{2}{c}{\textbf{Single}} & \multicolumn{2}{c}{\textbf{Merged}} & \multicolumn{2}{c}{\textbf{Context}} \\
         \cmidrule{2-7}
                         & F1 [\%] & $\sigma$ & F1 [\%] &  $\sigma$  & F1 [\%] &  $\sigma$ \\
         \midrule
         CoNLL 2003    & 96.27 & \small 0.15 & 96.01 & \small 0.31 & \textbf{96.79} & \small 0.26 \\
         GermEval 2014 & 87.18 & \small 0.18 & 85.91 & \small 0.62 & \textbf{87.48} & \small 0.40 \\
         CNEC 2.0      & 83.85 & \small 0.18 & 80.85 & \small 0.51 & \textbf{84.88} & \small 0.09 \\
         NKJP          & 92.21 & \small 0.14  & 91.41 & \small 0.17 & \textbf{92.32} & \small 0.11 \\
         KPWr (n82)    & 40.71 & \small 34.78 & 61.45 & \small 1.60 & \textbf{77.93} & \small 0.26 \\
         \bottomrule
    \end{tabular}
    \label{tab:models_strategy_eval}
\end{table}

In Table~\ref{tab:models_strategy_eval}, we can see that the models trained with the context representation perform better than the others. However, if we evaluate the model trained with the context representation using the single representation, we obtain significantly worse results (see Table~\ref{tab:models_strategy_eval_context_single}). This change in inference simulates a different scenario where the model is used to process short text snippets rather than long text fragments. We can observe a drop in performance from 3.50 pp for NKJP to 29.37 pp for CoNLL 2003. We also observe a much larger variance in the results, with the standard deviation varying from 1.20 to 40.28.

\begin{table}[ht]
    \caption{An average of F1 from 5 runs on the tune parts of the datasets using different setups on training and inference -- context for training and single for inference.}
    \centering
    \begin{tabular}{lccccc}
        \toprule
         \textbf{Training/Inference}   & \multicolumn{2}{c}{\textbf{Context/Context}} & \multicolumn{2}{c}{\textbf{Context/Single}} & \textbf{Diff} \\         
         \cmidrule{2-6}
         \textbf{Corpus}  & F1 [\%] &  $\sigma$  & F1 [\%] &  $\sigma$ & F1 [\%] \\
         \midrule
         CoNLL 2003    & \textbf{96.79} & \small 0.26 & 67.42 & \small 40.28 & -29.37 \\
         GermEval 2014 & \textbf{87.48} & \small 0.40 & 72.92 & \small 6.91  & -14.56 \\
         CNEC 2.0      & \textbf{84.88} & \small 0.09 & 69.49 & \small 4.16  & -15.39 \\
         NKJP          & \textbf{92.32} & \small 0.11 & 88.82 & \small 1.20  & -3.50\\
         KPWr (n82)    & \textbf{77.93} & \small 0.26 & 50.44 & \small 8.82  & -27.39\\
         \bottomrule
    \end{tabular}
    \label{tab:models_strategy_eval_context_single}
\end{table}

It can be argued that the problem lies in the absence of context as the representation of context provides more information than a single sentence. However, if we compare the results for the strategies of 'Context/Single' (as shown in Table~\ref{tab:models_strategy_eval_context_single}) with 'Single' (as shown in Table~\ref{tab:models_strategy_eval}), we can observe a similar drop in performance (see Table~\ref{tab:models_strategy_eval_context_single_2} for the side-by-side comparison). In fact, for four corpora, the results were significantly lower. The only exception was KPWr (n82), which showed an improvement of 9.73. Also, the variance for the Single/Single model was 34.78, indicating its instability.

\begin{table}[h!]
    \caption{An average of F1 from 5 runs on the tune parts of the datasets using different setups on training and inference -- context for training and single for inference.}
    \centering
    \begin{tabular}{lccccc}
    \toprule
         \textbf{Training/Inference}   & \multicolumn{2}{c}{\textbf{Single/Single}} & \multicolumn{2}{c}{\textbf{Context/Single}} & \textbf{Diff} \\         
         \cmidrule{2-6}
         \textbf{Corpus}  & F1 [\%] &  $\sigma$          & F1 [\%] &  $\sigma$ & F1 [\%] \\
         \midrule
         CoNLL 2003    & \textbf{96.27} & \small 0.15  & 67.42 & \small 40.28 & -28.85 \\
         GermEval 2014 & \textbf{87.18} & \small 0.18  & 72.92 & \small 6.91  & -14.26 \\
         CNEC 2.0      & \textbf{83.85} & \small 0.18  & 69.49 & \small 4.16  & -14.36 \\
         NKJP          & \textbf{92.21} & \small 0.14  & 88.82 & \small 1.20  & -3.39\\
         KPWr (n82)    & 40.71 & \small 34.78 & \textbf{50.44} & \small 8.82  & +9.73\\
         \bottomrule
    \end{tabular}
    \label{tab:models_strategy_eval_context_single_2}
\end{table}

\subsection{Observations}
\label{sec:observations}

After conducting our initial research, we identified two key observations:

\begin{enumerate}
    \item A model that is trained solely on either a dense (context) or sparse (sentence/merged) representation may not perform well on the opposite representation. The extent of this discrepancy varies based on the dataset.
    \item There is no single strategy that is stable across all tested corpora and dense/sparse representations. 
\end{enumerate}

We believe that these two limitations can be addressed by combining all strategies during the training process. This will result in a more stable model that can yield better performance regardless of the input strategy during inference. The results of our research are presented in the following section.

\section{Combined data representation for training}

Our proposed training procedure utilizes all three data representation strategies presented in Section~\ref{sec:representations}, which we call "union". The idea behind this approach is that a model trained on a specific representation tends to be biased toward that representation. In Section~\ref{sec:discrapency}, we demonstrated that a model trained on data with context performs poorly on data without context, while a model trained on data without context achieves good performance on data without context. By combining all three strategies, our approach performs a form of data augmentation. Each sample is used three times, but with slightly different contexts, which impacts the vector representation of each word due to the multi-head attention mechanism. As a result, the model becomes more resilient to context modifications. 

When it comes to training with union representation, we combine all other representations by putting together all the vectors for each representation. For instance, if a text is transformed into five vectors for a single representation, three vectors for a merged representation, and six vectors for a context representation, then for the union representation, we would use 14 vectors.

\begin{table}[ht]
    \caption{Comparison of the training strategies --- best F-measure for any {single}, {merged}, {context} vs. F-measure for {union} on the test subsets.}
    \centering
    \begin{tabular}{l|ccc|ccc}
        \toprule
        \textbf{Dataset} & \multicolumn{3}{c|}{\textbf{Tune}} &  \multicolumn{3}{c}{\textbf{Test}} \\
        \cmidrule{2-7}
                & \textbf{Best of } & \textbf{{Union}} & \textbf{Diff} & \textbf{Best of} & \textbf{{Union}} & \textbf{Diff}\\
                       & \small {single}, {merged},  & & & \small {single}, {merged}, \\
                       & \small {context}                   & & & {context} \\ 
        \midrule
         CoNLL 2003    & 96.79 & 96.95 & +0.16 & 93.54 & 93.69 & +0.15 \\
         GermEval 2014 & 87.48 & 88.13 & +0.65 & 87.57 & 88.04 & +0.57 \\
         CNEC 2.0      & 84.88 & 86.31 & +1.43 & 82.70 & 83.82 & +1.12 \\
         NKJP          & 92.32 & 92.78 & +0.46 & - & - & - \\
         KPWr (n82)    & 77.67 & 80.37 & +2.70 & 76.80 & 79.45 & +2.65 \\
         \bottomrule
         \end{tabular}
    \label{tab:diff_union}
\end{table}

In Appendix~\ref{sec:app1}, we have included Tables~\ref{tab:conll-tune-test} to~\ref{tab:kpwr-tune-test} that present detailed results for each combination of data representation during training and inference. We have considered three strategies for training --- {single}, {merged}, and {context} --- along with the combined strategy ({union}). For inference, we have included {single}, {merged}, and {context}, but left out the {union} as it's not suitable for inference. The last column of each row contains the average performance for all three strategies. We have reported the results for the tune and test subsets for each dataset, except for NKJP.

Based on the averaged F1 score values from five runs, Table~\ref{tab:diff_union} shows that the union representation strategy achieved the highest F1 score results across all datasets and for both the tuning and testing subsets. The results are compared with the best scores from other strategies such as single, merged, or context. The improvement in F-score ranges from +0.16/+0.15 for CoNLL 2003 dataset to +2.70/+2.65 for KPWr dataset.

Our main concern was the model stability for different data representations on inference. To verify the impact of union representation on model stability, we compared the averaged F1 score obtained for various data representations. Table~\ref{tab:diff_union_avg} compares the average F1 score for the models trained with context representation and the union representation. We compare our approach with the context representation as it is the best representation in the literature \cite{devlin2019bert,luoma-pyysalo-2020-exploring,wang-etal-2021-improving}. The comparison shows that the models trained with union data representation performed better across all datasets. The difference was from +0.73 for the NKJP dataset to +10.14/+9.98 for the CoNLL 2003 dataset and +11.65/+11.17 for the KPWr dataset. 

\begin{table}[ht]
    \caption{Comparison of the training strategies --- best average F-measure for any {single}, {merged}, {context} vs. average F-measure for {union} on the test subsets.}
    \centering
    \begin{tabular}{l|ccc|ccc}
        \toprule
        \textbf{Dataset} & \multicolumn{3}{c|}{\textbf{Tune}} &  \multicolumn{3}{c}{\textbf{Test}} \\
        \cmidrule{2-7}
          & \small \textbf{Avg.} &\small \textbf{Avg.} & \textbf{Diff} 
          & \small \textbf{Avg.} &\small \textbf{Avg.} & \textbf{Diff} \\
                       &\small context &  {union} & &\small context & {union} \\
        \midrule
         CoNLL 2003    & 86.52 & 96.66 & +10.14 & 83.00 & 92.98 & +9.98 \\
         GermEval 2014 & 82.24 & 88.05 &  +5.81 & 82.12 & 87.77 & +5.65 \\
         CNEC 2.0      & 79.22 & 85.84 &  +6.62 & 77.56 & 83.69 & +6.13 \\
         NKJP          & 91.07 & 92.77 &  +0.73 & - & - & - \\
         KPWr (n82)    & 68.32 & 79.97 & +11.65 & 67.63 & 78.80 & +11.17 \\
         \bottomrule
         \end{tabular}
    \label{tab:diff_union_avg}
\end{table}

We have noticed a significant difference in the level of improvement between the NKJP dataset and the next lowest improvement, which was +0.73 compared to +5.81. We believe that the low impact on the NKJP dataset could be correlated with its size, as it has almost four times more tokens and annotations than the second largest GermEval 2014 dataset.

\section{Comparison with SOTA}

In this section, we will be comparing the performance of the union strategy with the state-of-the-art (SOTA) models. The results for all datasets are summarized in Table~\ref{tab:sota}. Except for the CoNLL 2003 dataset, our results outperformed the SOTA. 

For the CoNLL 2003 dataset, our average F1 score was 93.69, which is 0.91 pp lower than the SOTA. However, we cannot compete with Wang, et al. (2020) \cite{DBLP:journals/corr/abs-2010-05006} as they used several pre-trained models to get word representation while we only used one pre-trained language model in our research.

\begin{table}[ht]
    \caption{Comparison of the {union} strategy with the SOTA models. {union} is an average from five runs.}
    \centering
    \begin{tabular}{lccclp{12em}}
        \toprule
        
        \textbf{Dataset} & Metric & \textbf{{union}} & \textbf{SOTA} & \textbf{Reference} & \textbf{Method}\\
        \midrule
         CoNLL 2003    & F1 & 93.69 & \textbf{94.60} & \cite{DBLP:journals/corr/abs-2010-05006} & Concatenation of several language models and context-based represenatation\\
         GermEval 2014 & F1 & \textbf{88.04} & 87.69 & \cite{MARCINCZUK2021291} & xlm-roberta large model with context-based representation \\
         CNEC 2.0      & F1 & \textbf{83.82} & 83.44 & Web page\footnote{\small\url{http://ufal.mff.cuni.cz/nametag/2/models}} & BERT and Flair embeddings with context-based representation \\
         NKJP         & Score & \textbf{92.60}* & 88.06 & \cite{Dadas2020}  & Neural iterative model based on ELMo language model\\
         KPWr (n82)    & F1 & \textbf{79.45} & 73.28 & \cite{Marcinczuk2019} & xlm-roberta large model with context-based representation\\
         \bottomrule
         \end{tabular}
    \label{tab:sota}
\end{table}

We conducted a straightforward comparison of GermEval 2014, CNEC 2.0, and KPWr (n82), using a single pre-trained model --- a larger version of XLM-Roberta that we also used in our research. Our analysis revealed that we achieved slight improvements for all three datasets, with GermEval 2014 seeing a 0.35 pp increase, CNEC 2.0 seeing a 0.38 pp increase, and KPWr (n82) experiencing a significant 5.83 pp improvement.

To evaluate the effectiveness of our approach for the NKJP dataset, we compared it to the SOTA results using the PolEval 2018 NER dataset\footnote{\url{http://2018.poleval.pl/index.php/tasks}} as a benchmark. The evaluation metric used was a weighted means of overlap and exact matches. Our score of 92.60 was significantly better than that of the SOTA model, which was 88.06 as reported in Table~\ref{tab:diff_union}. The improvement we achieved was 4.54 pp.

\section{Conclusions}

Based on our research, we have found that using context-based data representation can lead to instability in the inference process while dealing with different data representations. In simpler terms, when a named entity recognition model is trained on long text passages, it may not perform well when processing short text fragments. We have observed that this can result in a drop in performance of up to 29.37\%. However, a model trained on short fragments using a single sentence representation retains up to 99\% of the performance of the context-based model. This confirms that the decrease in performance is not due to the lack of context, but rather to the text representation bias.

To address this issue, we propose a training technique called union that combines all three data representation strategies (single, merged, and context). Union is a form of data augmentation that better utilizes the training data. Models trained with the union data representation are more resistant to the size of the input text. Both short fragments (single sentences) and long fragments (sentences with context) are processed with near the same performance. Our experimental verification showed that using this technique, we were able to improve the F1 score up to 11 pp. We also observed a correlation between the improvement and the size of the dataset. The smaller the dataset, the higher the improvement we can obtain. Conversely, the larger the dataset, the smaller the improvement can be. For example, for the NKJP dataset that is four times larger than the second-largest dataset, the improvement of F1 was only 0.73 pp.

In our experiments, we compared our model with the state-of-the-art models for the dataset we used. The results showed that the union data representation can significantly better utilize the training dataset. However, it cannot compete with another approach that uses multiple language models or external datasets. The SOTA model for the CoNLL 2003 dataset uses several language models by concatenating the word embeddings, while our approach based on one language model was 0.91 percentage points worse. For the remaining dataset, where the SOTA models are based on a single language model with context-based representation, we were able to improve the performance by up to 5.83 percentage points.

\bibliographystyle{unsrt}  


\clearpage

\setcounter{section}{0}

\appendix
\section{Detailed results for each dataset}
\label{sec:app1}

\begin{table}[h]
    \caption{The results of the evaluation on the CoNLL 2004 dataset. The model was trained on the train subset. The reported values are the average F1 scores obtained from five runs with different seeds}
    \centering
        \begin{tabular}{cp{5em}ccccccc}
            \toprule
               & & \multicolumn{6}{c}{Inference} \\
            \cmidrule{3-8}
                & & \multicolumn{2}{c}{\textbf{single}} & \multicolumn{2}{c}{\textbf{merged}} & \multicolumn{2}{c}{\textbf{context}} & \textbf{AVG} \\
        
             \cmidrule{3-8}
             Subset & Training  & F1  & $\sigma$ & F1 & $\sigma$ & F1 & $\sigma$ \\
             \midrule
             Dev & \textbf{single}         & \textbf{96.27} & \small  0.15 & 94.59          & \small 0.23 & 95.33 & \small 0.31 & 95.39 \\
                 & \textbf{merged}         & 68.72          & \small  4.02 & \textbf{96.01} & \small 0.31 & 95.87 & \small 0.12 & 86.87 \\
                 & \textbf{context}        & 67.42          & \small 40.28 & 95.36          & \small 0.49 & \textbf{96.79} & \small 0.26 & 86.52 \\
             \cmidrule{2-9}
                 & \textbf{union}          & \textbf{96.39} & \small 0.12 & \textbf{96.64} & \small 0.16 & \textbf{96.95} & \small 0.16 & \textbf{96.66} \\
             \midrule
             Test & \textbf{single}         & \textbf{92.43} & \small  0.17 & 90.82          & \small 0.62 & 91.79 & \small 0.59 & 91.68 \\
                  & \textbf{merged}         & 60.50          & \small  3.76 & \textbf{93.03} & \small 0.35 & 92.76 & \small 0.43 & 82.10 \\
                  & \textbf{context}        & 63.44          & \small 38.44 & 92.02          & \small 0.35 & \textbf{93.54} & \small 0.29 & 83.00 \\
             \cmidrule{2-9}
                  & \textbf{union}          & \textbf{92.18} & \small 0.29 & \textbf{93.06} & 0.22 & \textbf{93.69} & 0.14 & \textbf{92.98} \\
             \bottomrule
        \end{tabular}
\label{tab:conll-tune-test}
\end{table}

\begin{table}[h]
    \caption{The results of the evaluation on the GermEval 2014 dataset. The model was trained on the train subset. The reported values are the average F1 scores obtained from five runs with different seeds}
    \centering
        \begin{tabular}{cp{5em}ccccccc}
            \toprule
               & & \multicolumn{6}{c}{Inference} \\
            \cmidrule{3-8}
                & & \multicolumn{2}{c}{\textbf{single}} & \multicolumn{2}{c}{\textbf{merged}} & \multicolumn{2}{c}{\textbf{context}} & \textbf{AVG}\\
        
             \cmidrule{3-8}
             Subset & Training  & F1  & $\sigma$ & F1 & $\sigma$ & F1 & $\sigma$ \\
             \midrule
             Dev & \textbf{single}         & \textbf{87.18} & \small  0.18 & 84.59          & \small 0.90 & 85.76 & \small 0.79 & 85.84\\
                 & \textbf{merged}         & 57.85          & \small  6.56 & \textbf{85.91} & \small 0.62 & 85.84 & \small 0.77 & 76.53 \\
                 & \textbf{context}        & 72.92          & \small  9.91 & \textbf{86.32}  & \small 0.51 & {85.84} & \small 0.40 & 82.24\\
             \cmidrule{2-9}
                 & \textbf{union}          & \textbf{87.97} & \small 0.18 & \textbf{88.06} & \small 0.08 & \textbf{88.13} & \small 0.50 & \textbf{88.05}\\
             \midrule
             Test & \textbf{single}         & \textbf{86.95} & \small  0.30 & 84.09          & \small 0.56 & 85.57 & \small 0.35 & 85.54 \\
                  & \textbf{merged}         & 57.65          & \small  5.75 & \textbf{85.67} & \small 0.27 & 85.96 & \small 0.46 & 76.43 \\
                  & \textbf{context}        & 72.99          & \small  5.93 & 85.82          & \small 0.30 & \textbf{87.57} & \small 0.38 & 82.12\\
             \cmidrule{2-9}
                  & \textbf{union}          & \textbf{87.80} & \small 0.34 & \textbf{87.47} & 0.20 & \textbf{88.04} & 0.27 & \textbf{87.77}\\
             \bottomrule
        \end{tabular}
\label{tab:germeval-tune-test}
\end{table}

\begin{table}[h]
    \caption{The results of the evaluation on the CNEC 2.0 dataset. The model was trained on the train subset. The reported values are the average F1 scores obtained from five runs with different seeds}
    \centering
        \begin{tabular}{cp{5em}ccccccc}
            \toprule
               & & \multicolumn{6}{c}{Inference} \\
            \cmidrule{3-8}
                & & \multicolumn{2}{c}{\textbf{single}} & \multicolumn{2}{c}{\textbf{merged}} & \multicolumn{2}{c}{\textbf{context}} & \textbf{AVG}\\
        
             \cmidrule{3-8}
             Subset & Training  & F1  & $\sigma$ & F1 & $\sigma$ & F1 & $\sigma$ \\
             \midrule
             Dev & \textbf{single}         & \textbf{83.85} & \small  0.18 & 78.77          & \small 4.84 & 81.86 & \small 1.72 & 81.49\\
                 & \textbf{merged}         & 60.42          & \small  2.93 & \textbf{80.85} & \small 0.51 & 80.83 & \small 0.40 & 74.03 \\
                 & \textbf{context}        & 69.49          & \small  4.16 & 83.30  & \small 0.62 & \textbf{84.88} & \small 0.09 & 79.22\\
             \cmidrule{2-9}
                 & \textbf{union}          & \textbf{85.53} & \small 0.18 & \textbf{85.67} & \small 0.15 & \textbf{86.31} & \small 0.15 & \textbf{85.84}\\
             \midrule
             Test & \textbf{single}         & \textbf{81.98} & \small  0.30 & 76.73          & \small 3.89 & 79.83 & \small 1.64 & 79.51 \\
                  & \textbf{merged}         & 60.01          & \small  2.26 & \textbf{79.05} & \small 0.36 & 79.31 & \small 0.16 & 72.79 \\
                  & \textbf{context}        & 69.53          & \small  3.24 & 80.44          & \small 0.61 & \textbf{82.70} & \small 0.19 & 77.56\\
             \cmidrule{2-9}
                  & \textbf{union}          & \textbf{83.79} & \small 0.21 & \textbf{83.44} & 0.29 & \textbf{83.82} & 0.43 & \textbf{83.69}\\
             \bottomrule
        \end{tabular}
\label{tab:cnec-tune-test}
\end{table}

\begin{table}[h]
    \caption{The results of the evaluation on the NKJP dataset. The model was trained on the train subset. The reported values are the average F1 scores obtained from five runs with different seeds}
    \centering
        \begin{tabular}{cp{5em}ccccccc}
            \toprule
               & & \multicolumn{6}{c}{Inference} \\
            \cmidrule{3-8}
                & & \multicolumn{2}{c}{\textbf{single}} & \multicolumn{2}{c}{\textbf{merged}} & \multicolumn{2}{c}{\textbf{context}} & \textbf{AVG}\\
        
             \cmidrule{3-8}
             Subset & Training  & F1  & $\sigma$ & F1 & $\sigma$ & F1 & $\sigma$ \\
             \midrule
             Dev  & \textbf{single}         & \textbf{92.21} & \small  0.14 & 91.83          & \small 0.66 & 92.08 & \small 0.58 & 92.04 \\
                  & \textbf{merged}         & 3.40           & \small  1.72 & {91.41} & \small 0.17 & \textbf{91.52} & \small 0.11 & 62.11 \\
                  & \textbf{context}        & 88.82          & \small  1.20 & 92.06          & \small 0.23 & \textbf{92.32} & \small 0.11 & 91.07\\
             \cmidrule{2-9}
                  & \textbf{union}          & \textbf{92.63} & \small 0.45 & \textbf{92.66} & 0.10 & \textbf{92.78} & 0.13 & \textbf{92.77}\\
             \bottomrule
        \end{tabular}
\label{tab:nkjp-test}
\end{table}

\begin{table}[h]
    \caption{The results of the evaluation on the KPWr dataset. The model was trained on the train subset. The reported values are the average F1 scores obtained from five runs with different seeds}
    \centering
        \begin{tabular}{cp{5em}ccccccc}
            \toprule
               & & \multicolumn{6}{c}{Inference} \\
            \cmidrule{3-8}
                & & \multicolumn{2}{c}{\textbf{single}} & \multicolumn{2}{c}{\textbf{merged}} & \multicolumn{2}{c}{\textbf{context}} & \textbf{AVG}\\
        
             \cmidrule{3-8}
             Subset & Training  & F1  & $\sigma$ & F1 & $\sigma$ & F1 & $\sigma$ \\
             \midrule
             Dev & \textbf{single}         & \textbf{40.72} & \small  34.78 & 8.31          & \small 19.79 & 10.48 & \small 23.14 & 19.84 \\
                 & \textbf{merged}         &  6.38          & \small  1.86 & \textbf{61.45} & \small 1.60 & 61.12 & \small 1.37 & 42.98 \\
                 & \textbf{context}        & 50.44          & \small  7.82 & 76.85  & \small 0.32 & \textbf{77.67} & \small 0.26 & 68.32\\
             \cmidrule{2-9}
                 & \textbf{union}          & \textbf{79.28} & \small 0.50 & \textbf{80.25} & \small 0.25 & \textbf{80.37} & \small 0.15 & \textbf{79.97}\\
             \midrule
             Test & \textbf{single}         & \textbf{40.38} & \small  35.54 & 8.64          & \small 20.37 & 11.12 & \small 23.46 & 20.05 \\
                  & \textbf{merged}         & 5.70           & \small  1.23 & \textbf{61.00} & \small 1.53 & 60.36 & \small 1.70 & 42.36 \\
                  & \textbf{context}        & 50.33          & \small  7.74 & 75.75          & \small 0.63 & \textbf{76.80} & \small 0.65 & 67.63\\
             \cmidrule{2-9}
                  & \textbf{union}          & \textbf{77.89} & \small 0.18 & \textbf{79.06} & 0.21 & \textbf{79.45} & 0.15 & \textbf{78.80}\\
             \bottomrule
        \end{tabular}
\label{tab:kpwr-tune-test}
\end{table}

\end{document}